\documentclass{article} %
\usepackage[preprint]{colm2025_conference}

\usepackage{hyperref}
\usepackage{url}

\usepackage{lineno}

\definecolor{darkblue}{rgb}{0, 0, 0.5}
\hypersetup{colorlinks=true, citecolor=darkblue, linkcolor=darkblue, urlcolor=darkblue}

\usepackage{amsmath,amsfonts,bm}

\def\eqref#1{equation~\ref{#1}}
\def\1{\bm{1}}

\DeclareMathAlphabet{\mathsfit}{\encodingdefault}{\sfdefault}{m}{sl}
\SetMathAlphabet{\mathsfit}{bold}{\encodingdefault}{\sfdefault}{bx}{n}

\usepackage[utf8]{inputenc} %
\usepackage[T1]{fontenc}    %
\usepackage{booktabs}       %
\usepackage{amsfonts}       %
\usepackage{nicefrac}       %
\usepackage{microtype}      %
\usepackage{graphicx}
\usepackage{natbib}
\usepackage{doi}
\usepackage{amssymb}
\usepackage{amsmath}
\usepackage{appendix}
\usepackage{algorithm}
\usepackage{algpseudocode}
\usepackage{import}
\usepackage{cleveref}
\usepackage{pifont}
\usepackage{mdframed}
\usepackage{xcolor}
\usepackage{tcolorbox}
\usepackage{tabularx}
\usepackage{longtable}
\usepackage[greek,english]{babel}

\tcbuselibrary{skins}

\newtcolorbox{importantbox}[1][Important Note]{
    enhanced,
    colback=blue!5,
    colframe=blue!75!black,
    arc=4mm,
    boxrule=1.5pt,
    title=#1,
    fonttitle=\bfseries,
    attach boxed title to top left={yshift=-3mm, xshift=5mm},
    coltitle=black,
    boxed title style={
        enhanced,
        colback=blue!40,
        arc=1mm
    },
    top=15pt,                    %
    bottom=10pt
}

\newtcolorbox{valuesbox}[1][Value]{
    enhanced,
    colback=purple!5,
    colframe=purple!75!black,
    arc=4mm,
    boxrule=1.5pt,
    title=#1,
    fonttitle=\bfseries,
    attach boxed title to top left={yshift=-3mm, xshift=5mm},
    coltitle=black,
    boxed title style={
        enhanced,
        colback=purple!40,
        arc=1mm
    },
    top=12pt,                    %
    bottom=8pt
}

\definecolor{empirical}{RGB}{230, 159, 0} %
\definecolor{theoretical}{RGB}{0, 114, 178} %
\newcommand{\greencheck}{\textcolor{green!50!black}{\ding{51}}}  %
\newcommand{\orangecircle}{\textcolor{orange}{\ding{108}}}         %
\newcommand{\redcross}{\textcolor{red}{\ding{55}}}            %
\newcommand{\staricon}{\textcolor{yellow!80!black}{\ding{72}}}  %
\newcommand{\exclamationicon}{\textcolor{red}{\textbf{!}}}         %

\title{Evaluating Explanations: An Explanatory Virtues Framework for Mechanistic Interpretability \\
{\large The Strange Science: Part I.ii}}

\author{Kola Ayonrinde%
        \setcounter{footnote}{1}
        \thanks{
            Correspondence to:
            \texttt{koayon@gmail.com},
            \texttt{louis.yodj@gmail.com}
        } \\
        UK AI Security Institute
        \And
        {Louis Jaburi \textnormal{\textsuperscript{\textdagger}}}
}

\newcommand{\Prob}{\mathbb{P}}
\newcommand{\Ex}{\mathbb{E}}
\newcommand{\x}{\mathbf{x}}
\newcommand{\z}{\mathbf{z}}

\begin{document}

\maketitle

\begin{abstract}
    Mechanistic Interpretability (MI) aims to understand neural networks through causal explanations.
    Though MI has many explanation-generating methods,
    progress has been limited by the lack of a universal approach to evaluating explanations.
    Here we analyse the fundamental question ``What makes a good explanation?''
    We introduce a pluralist \emph{Explanatory Virtues Framework} drawing on four
    perspectives from the Philosophy of Science—the Bayesian, Kuhnian, Deutschian,
    and Nomological—to systematically evaluate and improve explanations in MI.
    We find that Compact Proofs consider many explanatory virtues and
    are hence a promising approach.
    Fruitful research directions implied by our framework include
    (1) clearly defining explanatory \textbf{simplicity},
    (2) focusing on \textbf{unifying} explanations and
    (3) deriving \textbf{universal principles} for neural networks.
    Improved MI methods enhance
    our ability to
    monitor, predict, and steer AI systems.
\end{abstract}

\section{Introduction}
\label{sec:intro}

Mechanistic Interpretability is the study of producing causal, scientific explanations
of artificial neural networks.
Good explanations allow us to monitor and understand AI systems
as well as providing affordances for steering and debugging.
But what is a \emph{good explanation}?

\citet{wu2024groups_unified_compact_proofs} observe the following problem:
When analysing the same algorithmic task,
\citet{chughtai2023group_universality} and \citet{standergrokking} produced
what appeared to be two valid Mechanistic Interpretability (MI) explanations
of the same model.
Yet the mechanisms that they propose are mutually inconsistent.
Without systematic criteria for choosing between explanations,
it is difficult to give good epistemic reasons for declaring one
explanation to be the better one.
Without good reasons to choose, researchers may either suspend judgement or
resort to disparate and subjective preferences.

Similarly, \citet{interp_illusions_2021}, \citet{interp_illusion_generalization_2024},
and \citet{subspace_illusion} find that explanations can be misleading.
While generated explanations might \emph{seem} plausible initially, they may turn out to be incomplete
---- or worse, complete confabulations which do not correspond to the
model internals.
Such explanations give only the illusion of model understanding.
We would like a clear guide as to which explanations are likely to be
explanatorily faithful to the model internals \citep{phil_exp_ayonrinde_jaburi},
and conversely which explanations may be unfaithful, even if
seemingly plausible.\footnote{
    In this sense, explanatory evaluations are useful to avoid
    interpretability researchers fooling themselves with Interpretability Illusions.
    As \citet{Feynman_1974} put it:
    ``The first principle is that you must not fool yourself —
    and you are the easiest person to fool.''
}

Recent work has developed evaluation metrics for interpretability with respect to either specific methods \citep{saebench2024},
or specific synthetic tasks \citep{interpbench, tracrbench}.
However, there is not a unifying framework that allows us to compare different explanatory methods
across a wide variety of tasks.

To address this problem, we introduce the \textbf{Explanatory Virtues Framework},
which answers the question:
\emph{Given two competing explanatory theories, which should we prefer?}
Our framework draws from the Philosophy of Science, specifically the
\emph{Bayesian},  \emph{Kuhnian}, \emph{Deutschian}, \emph{Nomological}
accounts of explanation
and we apply their criteria for theory choice to MI methods.
We examine the qualities that we should, and do, seek in good explanations,
via theoretical analysis and case studies respectively.
Using our Explanatory Virtues Framework,
we analyse four Mechanistic Interpretability methods:
Clustering, Sparse Autoencoders (SAEs), Causal Circuit Analysis, and Compact Proofs.
We find that the following Explanatory Virtues are often neglected among current MI methods:
\emph{Simplicity}, \emph{Unification}, \emph{Co-Explanation},
and \emph{Nomological Principles}.
We hence suggest pursuing these virtues as promising research directions.

The Explanatory Virtues Framework provides a systematic approach
for evaluating MI methods
and increasing our understanding of AI systems.
Such understanding is useful for AI Safety, AI Ethics, and AI Cognitive Science
\citep{bengio2025ai_safety_report, anwar2024foundational, chalmers2025propositional},
as well as debugging and improving neural networks
\citep{lindsay_bau_interp, sharkey2025openproblemsmechanisticinterpretability,Amodei_2025_urgency_interpretability}.

\paragraph{Contributions.} Our contributions are as follows:
\begin{itemize}
    \setlength{\itemsep}{0pt}
    \item Firstly, we provide a unified account of the Explanatory Virtues in MI.
    This can be understood as an answer to
    the question ``What makes a good explanation?''.
    \item Secondly, we analyse and compare MI methods
    with respect to these virtues.
    \item Finally, we suggest new directions for developing MI explanations,
    beyond the current state of the art.
\end{itemize}

\paragraph{Paper Structure.} This paper is organised as follows:
In \Cref{sec:valid}, we outline a definition of valid explanations in Mechanistic Interpretability,
distinguishing MI from other interpretability paradigms.
In \Cref{sec:values}, we analyse reasons for choosing one explanation over another
and introduce the \emph{Explanatory Virtues Framework}.
In \Cref{sec:explanations_in_the_wild}, we provide a critical analysis of
MI methods with respect to these Explanatory Virtues.
We conclude in \Cref{sec:discussion}
with a discussion of methodological
frontiers in interpretability,
and highlight virtues that we believe to be helpful in
developing more reliable explanations in MI.

\paragraph{Series Structure.}
This paper is the second in a series titled
\emph{The Strange Science of Mechanistic Interpretability},
concerning the Philosophy of Mechanistic Interpretability. \\
See \citet{phil_exp_ayonrinde_jaburi} (Part I.i)
for a discussion of the Philosophical Foundations
of Mechanistic Interpretability as a practise and its limitations.
See also \citet{ayonrinde2025bidirectional_interp} (Part I.iii)
which proposes methods to empower humans by teaching humans
Machine Concepts.

\section{Valid Explanations in Mechanistic Interpretability}

\label{sec:valid}

\emph{Neural network interpretability} (henceforth just \emph{interpretability}) is the process
of understanding artificial neural networks using the scientific method.
In this paper we focus on \emph{Mechanistic Interpretability (MI)}.
Following \citet{phil_exp_ayonrinde_jaburi}, we distinguish Mechanistic Interpretability from
other forms of interpretability noting that Mechanistic Interpretability produces
Model-level, Ontic, Causal-Mechanistic, and Falsifiable explanations.

\subsection{Explanations in Mechanistic Interpretability}
\label{sec:what_is_an_explanation}

Good scientific explanations provide answers to \emph{why} questions.
Typically a scientific explanation will provide an answer to the question
``Why did the phenomenon occur?'' and a good explanation will
enable the listener to better comprehend the phenomenon.
Explanations aim at knowledge.
As compression and comprehension are closely linked \citep{wilkenfeld2019understanding_compression},
good explanations \emph{compress observations by exploiting regularities in data}.

Neural networks are classically viewed as black-box prediction machines \citep{lipton2018mythos_classical_interp}.
However, \citet{phil_exp_ayonrinde_jaburi}
describe an alternative \emph{Explanatory View} of Neural Networks,
emphasising that deep neural networks contain \emph{representations} and \emph{mechanisms} that
can be understood as providing implicit explanations for their behaviour.
As models learn to generalize, they develop internal structures that compress information about the world.
Good explanations should uncover these internal structures.

\subsection{Defining Mechanistic Interpretability}
\label{sec:mech_explanations}

Following \citet{olah2020zoom_in_circuits,olsson2022induction_heads},
\citet{phil_exp_ayonrinde_jaburi}
define Mechanistic Interpretability as follows\footnote{
    See \citet{phil_exp_ayonrinde_jaburi} for a more complete
    exposition.
    Also see \Cref{sec:example_explanation_types} for
    intuitive examples of Explanation Types.
}:

\begin{importantbox}[Technical Definition of Mechanistic Interpretability  \citep{phil_exp_ayonrinde_jaburi}]
\label{box:mi_definition}

    Interpretability explanations are \textbf{valid} Mechanistic Interpretability explanations
    if they are
    \textbf{Model-level}, \textbf{Ontic}, \textbf{Causal-Mechanistic}, and \textbf{Falsifiable}.

    \begin{itemize}
        \setlength{\itemsep}{0pt}
        \item \textbf{Model-level}: Explanations should focus on understanding the neural network
        and not the sampling method or other system-level properties
        \citep{arditi_2024_systems_interp,compound-ai-blog}.
        \item \textbf{Ontic}: Explanations should refer to real entities within the model
        \citep{causal-realism-salmon}.
        \item \textbf{Falsifiable}: Explanations should yield testable predictions
        \citep{popper_scientific_discovery}.
        \item \textbf{Causal-Mechanistic}: Explanations should identify a step by step continuous causal chain from cause to phenomena,
        rather than statistical correlations or general laws
        \citep{woodward2003-causal_explanation,salmon1989four_decades_scientific_explanation,Bechtel2005_mechanistic_explanation}.
    \end{itemize}

\end{importantbox}

\section{The Virtues of Good Explanations}
\label{sec:values}
``\emph{Given two competing explanatory theories, which should we prefer?}''
This is the question of \emph{Theory Choice}
\citep{Kuhn1981_theory_choice, Schindler_theoretical_virtues, Kuhn1962-strcture_of_sci_rev}.
To answer this question we may look at the properties of explanations.

In the sciences, including in the Strange Science of Interpretability,
there can be no complete list of sufficient justificatory
criteria for an explanation.
Explanatory theories cannot be proven true, only falsified \citep{popper_scientific_discovery}.
However, there \emph{are} truth-conducive properties of explanatory theories.
We refer to such truth-conducive properties of explanations as \textbf{Explanatory Virtues}.
Explanatory Virtues are properties that are reliable indicators of truth.
Hence Explanatory Virtues are good reasons to prefer one explanatory theory to another.

Whether a property is an Explanatory Virtue is a \emph{normatively} loaded;
we should epistemically prefer explanations which embody Explanatory Virtues
as such explanations are more likely to be true and the aim of scientific explanation
is to aim at truth.\footnote{
    \citet{Schindler_theoretical_virtues} provides a
    discussion of the truth-conduciveness of the virtues we discuss.
}
Conversely, we \emph{descriptively} refer to properties of explanations that scientists value in practise
as \textbf{Explanatory Values}.

We can improve our explanatory theories by
increasing the virtue of our explanations.
Any individual theory may not embody all the explanatory virtues
but, all else equal, we ought to prefer theories that embody more explanatory virtues.
Similarly, we can improve the epistemic virtue of the
Mechanistic Interpretability scientific community by
looking to align our Explanatory Values with the Explanatory Virtues
\citep{sosa_1991_knowledge_perspective_virtue};
that is, by coming to appropriately value what is good
(i.e., truth-conducive) about explanations.

In this section, we discuss Explanatory Virtues --
the properties that ML researchers \emph{should} value.
We assess four accounts of explanation:
the Kuhnian, Bayesian, Deutschian, and Nomological accounts.
If these accounts correctly identify properties that we ought to value,
then the combined set of such properties are Explanatory Virtues.
These properties will form our pluralist \textbf{Explanatory Virtues Framework}.
We provide a mathematical definition for each Explanatory Virtue
which serves to ensure that there is a consistent and canonical way to compute
each virtue thus allowing for a more objective comparison of
explanations.\footnote{
    We believe that canonical definitions are important because Mechanistic Interpretability
    has previously had several definitions which have been used inconsistently
    by different researchers, making it difficult to directly compare across methods.
    Readers may check their intuitive understanding of the textual
    definitions against the mathematical definitions to ensure they are consistent.
    We also include a rubric for assessing explanatory methods in \Cref{table:values_rubric}.
    We further include intuitive examples to illustrate the Explanatory Virtues
    in \Cref{sec:examples_for_explanatory_values}.
}
Then in \Cref{sec:explanations_in_the_wild}, we will discuss
what MI researchers \emph{do} value in practise,
that is the Explanatory Values in Mechanistic Interpretability.
We provide a summary of our Pluralist Explanatory Virtues Framework
and how the virtues relate to each other in \Cref{fig:explanatory_values}.

\paragraph{Notation.}
We denote the explanation under consideration as $E \in \mathcal{E}$,
where $\mathcal{E}$ is the set of all possible explanations
and $B$, the background theory.
$\x_T$ denotes observational data that the explanation is fitted to (training data).
We assume $x_T$ is sampled from the set of possible observational data $\mathcal{X}$.
$\x_I$ denotes future observational data that was not accessible at
explanation-making time (inference-time data).
$x_{T,i}$ is the $i$-th data point in $\x_T$.
We denote $k$ a complexity measure
(for example, Kolmogorov complexity) and $|E|_B$ the
description length of an explanation $E$ under background theory
$B$ measured in bits.

\subsection{Bayesian Theoretical Virtues}
\label{sec:bayes_values}

\citet{dedeo_2020_probability_consilience} describe a Bayesian approach to Inference to the Best Explanation
\citep{henderson2014bayesianism_inference_best_explanation}.
Here, the Explanatory Virtues are the credence-raising properties of the theory.
These virtues can be split into two categories:
\textcolor{blue}{\textbf{theoretical virtues}} (in blue),
which are properties of the explanation that do not
depend on any observed or yet to be observed data, and
\textcolor{orange}{\textbf{empirical virtues}} (in orange),
which are properties of the explanation that are
defined in relation to the observed data.

\paragraph{Accuracy, Precision, and Priors.}
The Bayesian virtues are the empirical Explanatory Virtue of \textcolor{orange}{\textbf{Accuracy}},
the theoretical Explanatory Virtue of \textcolor{blue}{\textbf{Precision}} and the \textcolor{blue}{\textbf{Prior}}
probability of some explanation given the background theory.

\textcolor{orange}{Accuracy} represents the probability of the true data given the explanation.
Log-likelihood is the logarithm of Accuracy.
Similarly, \textcolor{blue}{Precision} is the expected log-likelihood of data conditional on the explanation being true.
Precision represents the degree to which an explanation's predictions concentrate in a particular region
of the space of possible observed data. Higher precision means that the explanation is more constraining
in its predictions, making risky and useful predictions that rule out other possibilities,
if the explanation is correct.
\footnote{Note that the definition of Precision here is a slightly different notion
to the Precision metric in Machine Learning as in `Precision-Recall' analysis
\citep{hastie2009elements_statistical_learning}.
There, Precision is the fraction of true positives among the predicted positives.
Here, by Precision we mean to say that more precise explanations are more constraining in their predictions.}

We decompose \textcolor{orange}{Accuracy} and \textcolor{blue}{Precision} into
further Explanatory Virtues as follows.

\paragraph{Descriptiveness and Co-Explanation.}
Given many data points $\x = \{x_{1}, x_{2}, \ldots, x_{n}\}$, we would like to understand
how well an explanation explains each data point in isolation and how well it explains
multiple data points together.
We hence define \textcolor{orange}{\textbf{Descriptiveness}} as the component
of Log-Likelihood where data observation is considered in isolation and
\textcolor{orange}{\textbf{Co-Explanation}} as the component of Log-Likelihood which focuses on how an explanation
can explain multiple data points, above its ability to predict any single observation in isolation.

\paragraph{Power and Unification.}
Analogously, we can break down \textcolor{blue}{Precision} into our theoretical virtues of \textcolor{blue}{\textbf{Power}}
and \textcolor{blue}{\textbf{Unification}},
defined analogously where \textcolor{blue}{Power} measures the ability to explain individual data points
and \textcolor{blue}{Unification} measures the ability to
connect multiple disparate observations together.

\begin{valuesbox}[Glossary of Bayesian Virtues]
    \begin{align}
        \tag{Accuracy} \label{eq:accuracy}
        Acc(E) &= \Prob(\x_T | E) \\
        \tag{Precision} \label{eq:precision}
        Prec(E) &= \Ex_{x_T \sim \mathcal{X}}[\log(\Prob(\x_T | E))] \\
        \tag{Prior} \label{eq:prior}
        Prior(E) &= \Prob(E | B) \\
        \tag{Descriptiveness} \label{eq:descriptiveness}
        Desc(E) &= \sum_{i}\log(\Prob(x_{T,i} | E)) \\
        \tag{Co-explanation} \label{eq:co_explanation}
        CoEx(E) &= \log(Acc(E)) - Desc(E) = \log(\frac{\Prob(\x_T | E)}{\prod_{i}\Prob(x_{T,i} | E)}) \\
        \tag{Power} \label{eq:power}
        Power(E) &= \Ex_{x_T \sim \mathcal{X}}[\sum_{i}\log(\Prob(x_{T,i} | E))] \\
        \tag{Unification} \label{eq:unification}
        Unif(E) &= Prec(E) - Power(E) = \Ex_{x_T \sim \mathcal{X}}\log(\frac{\Prob(\x_T | E)}{\prod_{i}\Prob(x_{T,i} | E)})
    \end{align}

\end{valuesbox}

\subsection{Kuhnian Theoretical Virtues}
\label{sec:kuhn_values}

\citet{Kuhn1981_theory_choice} lists five theoretical virtues as
a basis for theory choice: \textcolor{orange}{\textbf{Accuracy}}, \textcolor{blue}{\textbf{(Internal) Consistency}},
\textcolor{blue}{\textbf{Scope (Unification)}}, \textcolor{blue}{\textbf{Simplicity}} and \textcolor{orange}{\textbf{Fruitfulness}}.
We previously explored \textcolor{blue}{Unification} (\textcolor{blue}{Scope}) and \textcolor{orange}{Accuracy} in \Cref{sec:bayes_values}.

\paragraph{Accuracy and Fruitfulness.} \textcolor{orange}{Accuracy} is the extent to which
the explanation fits the
available data at the time of the creation of such an explanation.
We can think of this as the ``mundane empirical success'' of an explanation,
which we can contrast with the ``\textcolor{orange}{novel empirical success}'' of an explanation or its
\textcolor{orange}{\textbf{Fruitfulness}} \citep{Lakatos1978_novel_success}.
Machine Learning researchers may draw a close analogy here with
\textcolor{orange}{Accuracy} being a performance measure on the training/validation set and
\textcolor{orange}{Fruitfulness} being a performance measure on a (naturally held-out) test
set.\footnote{
    Here we allow for the test set to be drawn from the same distribution as the training set or
    to represent a distribution shift.
    In science, the analogy of the training and test set being drawn from the same distribution, would be if
    the explanation also fits new data that we didn't observe before making the explanation but we plausibly could have observed.
    The analogy for the distribution shift would be if having the new explanation makes us adversarially
    seek out new observations to attempt to falsify our explanatory theory.
    In the latter case, we wouldn't plausibly have observed these new data points before making the explanation.
}
Fruitful explanations have reach:
they usefully generalise beyond the context of the original
problem that the explanation was designed to solve.

One particularly important type of Fruitfulness that interpretability researchers
often test MI methods for is \emph{Pragmatic Utility},
the ability for a local explanation to be useful on some
downstream task of interest \citep{Marks_2025_downstream_applications}.
For example, researchers have tested Sparse Auutoencoder features on downstream tasks
such as unlearning \citep{saebench2024}, probing \citet{axbench},
building robust classifiers \cite{openai_saes_2024},
and building sparse feature circuits \citet{marks_sparse_feature_circuits_2024}.
Analysing downstream pragmatic utility ensures that MI methods
are directly useful in aiding researchers in tasks that they
care about \citep{lindsay_bau_interp,Amodei_2025_urgency_interpretability}.

\paragraph{Consistency.} A necessary criterion for a theory to be a good explanation is that it is internally \textcolor{blue}{consistent}.
That is to say, the explanation must not contain any logical contradictions.

\paragraph{Simplicity.} \textcolor{blue}{\emph{Simplicity}}
is considered a key virtue for scientific explanations
\citep{White2005-simplicity, Qu2023-hume_simplicity, mackay2003information}.
However, there are many forms of \textcolor{blue}{simplicity} that may be chosen,
which may rank explanations differently \citep{Lakatos1970_simplicity_is_relative}.
We consider the main three forms of measures of simplicity:
\textcolor{blue}{\emph{Parsimony}}, \textcolor{blue}{\emph{Conciseness}} and \textcolor{blue}{\emph{Complexity}}.
\textcolor{blue}{\textbf{Parsimony}} counts the number of entities that are
posited by the explanation \citep{dedeo_2020_probability_consilience}.\footnote{
    Parsimony is slippery to define well in practise as it is not always clear what counts as an entity.
    Worse still, parsimony might treat intuitively highly complex objects and very simple objects both equivalently
    as ``entities'' and simply count them up without nuance.
    \citet{parsimony-baker} provides a discussion of the downsides of Parsimony as a measure of simplicity.
}
\textcolor{blue}{\textbf{Conciseness}} is a Shannon-complexity measure of the information in an explanation
given by the description length \citep{shannon1948communication,mackay2003information},
\textcolor{blue}{\textbf{(K-)Complexity}} is a Kolmogorov-complexity measure of an explanation in terms of the shortest
program that can generate it \citep{kolmogorov1965information,hutter2023aixi}.
For all simplicity measures, lower values are preferred.

\begin{valuesbox}[Glossary of Kuhnian Virtues]
    \begin{alignat}{2}
        \tag{Fruitfulness} \label{eq:fruitfulness}
        Fruit(E) &= && \mathbb{P}(\mathbf{x}_I|E) \\
        \tag{Consistency} \label{eq:consistency}
        \text{E is inconsistent} &\iff && E \vDash \bot \\
        \tag{Parsimony} \label{eq:parsimony}
        Pars(E) &= && \#\_\text{of}\_\text{entities}(E) \\
        \tag{Conciseness} \label{eq:conciseness}
        DL(E) &= && |E|_B \\
        \tag{Complexity} \label{eq:complexity}
        \text{k-Compl}(E) &= && k(E)
        \end{alignat}
\end{valuesbox}

\subsection{Deutschian Theoretical Virtues}
\label{sec:deutsch_values}

\paragraph{Falsifiability and Hard-to-Varyness.}\citet{popper_scientific_discovery} writes that the key criteria of science is that its theories
should be \textcolor{blue}{\textbf{Falsifiable}} - that is, our explanations should
come with a clear set of testable predictions attached.
\citet{deutsch2011beginning_of_infinity} further argues that alongside falsifiability,
we should also seek explanations which themselves are \textcolor{blue}{\textbf{Hard-To-Vary}}.
Intuitively we might think of an explanation E as \textcolor{blue}{hard-to-vary} if it cannot be
easily modified to account for incoming data that contradicts the explanation.
More precisely consider a modification $\Delta$ to an explanation E,
where $\Delta$ is some edit operation formed of a list of insertions,
deletions, substitutions and transpositions of symbols in E.
$|\Delta|$ is the number of such operations in $\Delta$.

The \textcolor{blue}{hard-to-varyness} criteria then captures the intuition that if you
add some modification or ``epicycle'' $\Delta$ to an explanation E,
then the new explanation E' should have lower novel empirical success than E (complexity-weighted).
Conversely, if we can add some modification to an explanation
and the new explanation has higher mundane and novel empirical success without
being more complex, then we should prefer the new explanation.\footnote{
    We provide a complementary adhocness metric in \Cref{sec:adhocness}.
}

For some complexity measure k, we can then say that an explanation E is \textcolor{blue}{hard-to-vary}
if it is at a local maximum
of the function $hv(E) = \log(Acc(E)) - k(E)$.\footnote{
    We informally consider two explanations close if they are a small number of edit operations apart.
}

\begin{valuesbox}[Hard-to-Varyness]
    An explanation $E$ is hard-to-vary if it is at a local maximum of the function
    \begin{equation}
        \label{eq:hard_to_vary}
        \tag{Hard-to-Varyness}
        hv(E) = \log(Acc(E)) - k(E)
    \end{equation}
\end{valuesbox}

\subsection{Nomological Theoretical Virtues}
\label{sec:nomological_values}

In \citet{hempel_explanation}'s Deductive-Nomological (DN) model of explanation,
a scientific explanation is a \emph{sound} \emph{deductive} argument where
at least one of the premises is a ``general law''.
For our purposes, we can think of general laws as
``for all'' statements which are true and not accidentally true.
General laws describe necessary rather than contingent facts of the world.
For example, ``all gases expand when heated under constant pressure'' is a general law
whereas ``all members of the Greensbury School Board for 1964 are bald'' might be true but only
by coincidence, as it were.

\paragraph{Nomologicity.} Though we do not require our explanations to precisely follow the DN model of explanation,
the \textcolor{blue}{\textbf{Nomologicity}} (or \textcolor{blue}{\emph{Lawfulness}}) of an explanation, i.e. whether
the explanation appeals to general laws or derives universal principles,
is an explanatory virtue.\footnote{
    \citet{myers2012cognitive_styles_nomological} contrasts nomological explanations with \emph{mechanistic} explanations,
    considering the former to be explanations at a higher level and the latter to be explanations at a lower level.
    However, \citet{phil_exp_ayonrinde_jaburi} note that this is a false dichotomy:
    The entities in mechanistic explanations can be emergent entities and citing general laws can
    aid in the explanation and unification of low-level phenomena.
}

\begin{valuesbox}[Nomologicity]
    An explanation $E$ is nomological if it appeals to general laws or universal principles
    about neural networks.
\end{valuesbox}

\subsection{Explanatory Virtues for Mechanistic Interpretability}
\label{sec:explanatory_virtues}

\begin{figure}[h]
    \centering
    \begin{minipage}{1.0\linewidth}
        \centering
        \includegraphics[width=\linewidth]{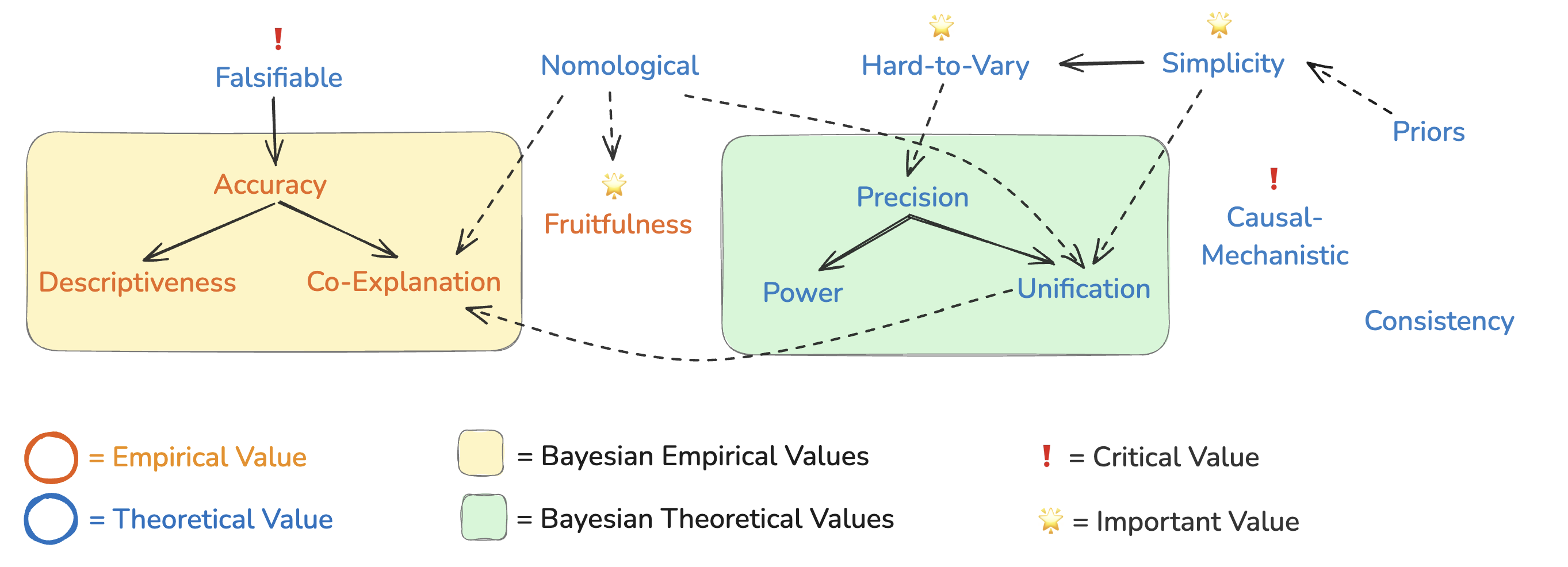}
    \end{minipage}
  \caption{A Directed Acyclic Graph representation of the \textbf{Explanatory Virtues Framework}
    showing the relationships between virtues.
    Empirical virtues are coloured orange and theoretical virtues are coloured blue.
    We show the virtues which directly depend on each other with bold arrows ($\boldsymbol{\rightarrow}$)
    and those which are highly related with dashed arrows ($\dashrightarrow$).
    The Explanatory Virtues which are essential for any scientific explanation
    (Falsifiability and Causal-Mechanisticity) to be valid are denoted with an exclamation mark;
    the most important virtues to decide between explanations (Simplicity, Hard-to-Varyness, and Fruitfulness)
    are marked with a star.
    \Cref{sec:values_rubric} details a rubric for assessing explanatory methods.
    \Cref{sec:straightforward_explanations} provides an example illustrating the importance
    of Simplicity as an explanatory virtue.
}
\label{fig:explanatory_values}
\end{figure}

We provide a summary of our pluralist Explanatory Virtues Framework
and how the virtues relate to each other in \Cref{fig:explanatory_values}.
These explanatory virtues are not necessarily exhaustive nor completely
independent of one another.\footnote{
    We detail an additional possible virtue in \Cref{sec:local_decodability}.
}
Some virtues may be in tension with each other.
For example, Accuracy may be traded off against
Simplicity in some cases.
Here we may aim to be at the optimal point of this trade-off on a
Pareto frontier.

We hope the reader may agree that our Explanatory Virtues both are
(1) important considerations for the evaluation of explanations and
(2) truth-conducive.\footnote{
    That is to say that
    all else equal explanations which embody these virtues are more likely to be true.
    We refer readers to \citet{Schindler_theoretical_virtues} for a detailed
    discussion of the truth-conduciveness of many of the virtues we discuss.
}
Thus, these virtues can be a useful guide for theory choice
and, more generally, can aid in the developments of new explanatory
methods.
Mechanistic Interpretability researchers, we argue,
ought to value the Explanatory Virtues.

For an explanation to be a \emph{good} explanation in Mechanistic Interpretability,
it must first be a \emph{valid} MI explanation.
In \Cref{sec:mech_explanations} we identified valid MI explanations as those
which are Model-Level, Ontic, Causal-Mechanistic, Falsifiable.
Validity requires all of the four validity conditions above to be met.
The Explanatory Virtues, then, allow us to assess the quality of
valid MI explanations and provide epistemic reasons to
prefer one explanation over another.

\section{Explanations in the Wild: Case Studies in Mechanistic Interpretability}
\label{sec:explanations_in_the_wild}

In \Cref{sec:values}, we explored the Explanatory Virtues.
These values included the Theoretical Explanatory Virtues of
\emph{Precision}, \emph{Power}, \emph{Unification}, \emph{Consistency},
\emph{Simplicity}, \emph{Nomologicity},
\emph{Falsifiability} and \emph{Hard-To-Varyness} as well as
the Empirical Explanatory Virtues of
(Mundane) \emph{Accuracy}, \emph{Descriptiveness}, \emph{Co-Explanation} and \emph{Fruitfulness}.
We now consider how these virtues are instantiated in
the methods that Mechanistic Interpretability researchers use
in practice.
That is, we consider how \emph{valued} each Explanatory Virtue is within MI methods.

Visual summaries of the methods we discuss
in this section can be found
in \Cref{sec:explanations_figures}.

\subsection{Examples}

\subsubsection{Clustering (Activations or Inputs)}
\label{sec:clustering}

One primitive form of neural network explanation is a clustering of
model inputs or activations.
For a complex model, such an explanation will not typically be highly accurate.
However, this explanation \emph{is} a simplification of the overall model performance.
Here we might imagine finding some partition of the input/activation space,
mapping a given input $\x$ to its associate cluster, of which $\x$ is ideally a typical member.
Then we may take the cluster (and possibly the output of the model on some cluster representative)
as a proxy for the model's behaviour.\footnote{
    We may think of the clustering explanation as performing
    some ``quotienting'' operation of the input space
    by the equivalence relation of being in the same cluster.
}

Though this explanation is clearly not sufficient in many cases, we note that
it does perform some compression of the input space and we can control the
simplicity of the explanation by varying the number of clusters.
Similarly, the explanation generated here is Falsifiable; we can test
how well our cluster model predicts the behaviour of the original model.
However, this explanation clearly falls down by not being \hyperref[sec:mech_explanations]{Causal-Mechanistic} in nature,
and the Fruitfulness of the explanation may be low if the procedure is vulnerable to
outliers.

\subsubsection{Sparse Autoencoder Explanations of Representations/Activations}
\label{sec:sae_explanations}

Sparse Autoencoders (SAEs) can be used to decompose the representations
of neural activations into a linear combination of sparsely activating,
disentangled and monosemantic latents \citep{anthropic_sae_towards_monosemanticity_bricken, huben2024saes}.
Though many evaluation schemes have been proposed for SAEs \citep{saebench2024,axbench},
the primary axes on which SAE explanations are evaluated is on \emph{Empirical accuracy}
and \emph{Simplicity}.
Here Accuracy represents either a local unsupervised accuracy measure like reconstruction error,
or the downstream performance of the interpreted model when the SAE reconstructions
are patched into the model in place of the original activations.

\paragraph{MDL-SAEs.}
\citet{ayonrinde2024_mdl_saes} provide a useful case study of how different types of
Simplicity measures may be more or less principled in different contexts.
Within the MDL-SAE framework, SAE explanations are evaluated on
Accuracy, Novel Empirical Success and Conciseness,
where \emph{Conciseness} is an information theoretic
measure of Simplicity (see \Cref{sec:kuhn_values}).
This stands in contrast to the classical SAE framework where the
simplicity measure is instead the SAE latent sparsity,
a \emph{parsimony} measure.
In this case, changing the simplicity measure from sparsity (Parsimony) to
description length (Conciseness) solved three key problems for SAEs:
avoiding undesired feature splitting, enabling principled choice of SAE width,
and ensuring uniqueness of feature-based explanation \citep{kayonrinde2025incoherent_saes}.

\paragraph{Explanatory Virtues for SAEs.}
SAE explanations, like most ML methods, value Falsifiability and Novel Empirical Success
(predictions beyond the training set). There is also some Co-Explanatory power
in that the same feature dictionary should be used to explain any activations
(at least from the same layer of the model).
However, SAE explanations might be Ad-hoc and not Hard-to-Vary.
As noted by \citet{apollo_end2end_sae}, contributions from features activated
on SAEs trained for reconstruction may have little effect on the
downstream performance of the model.
Hence the corresponding feature activations are effectively free parameters.
Similarly, the tendency to enlarge the feature dictionary (i.e.\ increase the SAE width)
or add additional active features to explanations (i.e.\ increase the allowable
$\ell^0$ norm of the feature activations vector)
without clear justification, suggests an implicit ad-hocness in the explanations.
MDL-SAEs provide some guidance against the ever increasing size of the feature dictionary,
however it still remains an open question as to how to ensure that SAE explanations are
truly hard-to-vary and pick out features which are causally relevant to the
downstream behaviour of the model \citep{leask2025meta_saes}.

\subsubsection{Causal Abstraction Explanations of Circuits}
\label{sec:circuits}

As in neuroscience, a natural way to explain the behaviour of a neural network
for interpretability researchers is to decompose the network into circuits
\citep{olah2020zoom_in_circuits, kandel2000principles_neuroscience}.
Circuits can be formally specified by a correspondence between the network and
some understood high-level causal model using the theory of Causal Abstractions
\citep{geiger2023causal,woodward2003-causal_explanation, Beckers_halpern_2019_Abstracting_Causal_Models, pearl2009causality}.
In particular, the notion of abstraction that is typically appealed to is constructive
abstraction \citep{Beckers_halpern_2019_Abstracting_Causal_Models}.
Paraphrasing from \citet{geiger_2021_causal_abstraction},
a high-level model (an understandable causal model) is a \emph{constructive abstraction}
of a low-level model if we can partition the variables in the low-level model
(e.g. the neural network neurons) such that:
\begin{enumerate}
  \setlength{\itemsep}{0pt}
  \item Each low-level partition cell can be assigned to a high-level variable.
  \item There is a systematic correspondence between interventions on the
low-level partition cells and interventions on the high-level variables.
\end{enumerate}

The Causal Abstraction framework for circuit analysis clearly focuses on the
Falsifiability of explanations and the \emph{Faithfulness} of the explanation
to the underlying causal model
(Empirical Accuracy and Novel Success under interventions).
To encourage simplicity in explanations,\footnote{
    After all, it's not clear what the point of a highly complex abstraction
    would be when the network itself can be viewed as a causal model if
    we disregarded the simplicity criterion.
},
we may also seek \emph{Completeness} and \emph{Minimality}
in circuit explanations \citep{wang_ioi}.
(Behavioural) Faithfulness, Completeness, and Minimality are denoted the \emph{FCM} criteria for
circuit explanations.

\paragraph{FCM Criteria for Circuits.}
For $C$ a proposed circuit and $M$ the model, the \textbf{Completeness} criterion states that
for every subset $K \subset C$,
the incompleteness score $|F(C \setminus K) - F(M \setminus K)|$ should be small.
Intuitively, a circuit is complete if the function of the circuit and the model
remain similar under ablations.
Conversely, the \textbf{Minimality} criterion states that for every node $v \in C$
there exists a subset $K \subseteq C \setminus \{v\}$ that has high minimality
score $|F(C \setminus (K \cup \{v\})) - F(C \setminus K)|$.
Intuitively, a circuit is minimal if it doesn't contain components which are unnecessary
for the function of the circuit.

Algorithms such as ACDC \citep{conmy2023acdc}
find circuits that (approximately) satisfy the FCM criteria.
However, it is well known \citep{wang_ioi} that the FCM criteria are in tension
and that it is not always possible to satisfy all three criteria simultaneously.
In practise, finding circuits is a computationally
challenging problem and circuit discovery algorithms typically
only find approximately optimal circuits
\citep{adolfi2024circuit_complexity}.\footnote{
    \citet{shi2024hypothesis_testing_circuits} provide
    hypothesis tests for circuits which test the related criteria of
    equivalence, independence and minimality.
    Their approach is a practical method for evaluating
    circuit explanations.
}

\paragraph{Explanatory Virtues for Circuit Explanations.}
Despite the virtues of these approaches, they however do suffer from
poor unification, co-explanation and nomologicity.
In both manual and automated circuit discovery methods,
most attention is paid to individual circuits rather than the
relation and composition of subcircuits.
Circuit explanations for two related tasks
which share internal components are not typically privileged.
Similarly, there are often no general laws or principles that detail
which circuits are likely to be found in a network,
and how these circuits relate to one another across contexts.

\subsection{Compact Proofs}
\label{sec:compact_proofs}

The above examples of Clustering,
SAEs and Circuits are methods for both the \emph{creation} of explanations
and also provide \emph{evaluation methods} for the explanations created.
The Compact Proofs methodology \citep{gross2024compact,wu2024groups_unified_compact_proofs, jaburi2025finetuning}
is a method for evaluating \emph{any} Causal-Mechanistic explanations obtained through other methods.
In the Compact Proofs framework, an explanation is converted into a formal guarantee
that allows researchers to assess the Accuracy and Simplicity of the explanation.

Given a data distribution $\mathcal{D}$,
and a model $M_\theta$ with weights $\theta \in \mathcal{W}$,
we would like to obtain a lower bound for the model's accuracy
over $\mathcal{D}$.\footnote{
    In general, we might be interested in bounding metrics which are to be
    minimised (e.g. loss) rather than maximised (e.g. accuracy and reward).
    In that case we may seek upper bounds rather than lower bounds but the argument is otherwise analogous.
}
Formally, we construct a \emph{verifier} program $V(\theta, E)$,where $E$ is the explanation.
The aim for $V$ is to return a worst-case bound on the model's
performance that is as tight as possible
with the proof that bound holds being as
computationally efficient as possible.
We may think of the computational efficiency as a
measure of the simplicity of the proof \citep{Xu2020v_information}.
Note that these two goals,
the \emph{tightness} (Accuracy) of the bound and the \emph{compactness} (Simplicity)
of the proof (explanation), are in tension with one another.
A good explanation should push out the (tightness, compactness)-Pareto
frontier.\footnote{
    \Cref{sec:straightforward_explanations} provides an example
    of one basic proof strategy which is computationally expensive but
    provides a tight bound.
    This strategy is known as the \emph{brute force proof} \citep{gross2024compact}
    and corresponds to the \emph{straightforward, Implementation-level explanation} \citep{phil_exp_ayonrinde_jaburi}.
}

\citet{gross2024compact} show that faithful mechanistic explanations lead to
tighter performance bounds and more efficient (i.e. simpler) proofs.
Informally, we may say that Compact Proofs allow us to leverage
good MI explanations into tighter and more compact proof bounds.
We note that this method allows for finding and evaluating explanations
which satisfy many of the Explanatory Virtues: Precise explanations allow
for tighter bounds, Accuracy and Simplicity are directly optimised for,
and Causal-Mechanistic explanations are generally required for
non-vacuous bounds.\footnote{
    At present it is not known how to scale the Compact Proofs methodology
    to much larger models with additional superposition noise and still get
    informative, non-vacuous bounds. This remains an open problem and a gold standard
    for evaluating explanations.
}

\subsection{Discussion of Explanatory Values}
\label{sec:values_discussion}

\Cref{table:values_case_study} shows that
some Explanatory Virtues are consistently valued
highly across different methods.
However, all current interpretability methods could be improved on some dimension
to be more likely to produce
human-understandable and useful explanations.
In particular, we suggest that methods which produce or appeal to
nomological principles and which unify accounts of neural
network behaviour are likely to be increasingly successful.

\begin{table}[H]
    \centering
    \begin{tabular}{lccccc}
    \toprule
    \textbf{Explanatory Virtue} & Importance & Clustering & (MDL) SAEs & Circuits & Compact Proofs \\
    \specialrule{1.5pt}{3pt}{3pt}
    \multicolumn{6}{l}{\hyperref[sec:mech_explanations]{\textit{Validity}}} \\
    {\textbf{\color{theoretical}Causal-}} &  \\
        \textbf{\color{theoretical}Mechanistic} & \exclamationicon & \redcross & \orangecircle & \greencheck & \greencheck \\
    \midrule
    \multicolumn{6}{l}{\hyperref[sec:bayes_values]{\textit{Bayesian}}} \\
    {\color{theoretical}Precision} &  & \orangecircle & \orangecircle & \greencheck & \greencheck \\
    {\color{theoretical}Priors} &  & \orangecircle & \orangecircle & \redcross & \redcross \\
    {\color{empirical}Descriptiveness} &  & \orangecircle & \orangecircle & \greencheck & \greencheck \\
    {\color{empirical}Co-explanation} &  & \redcross & \redcross & \redcross & \orangecircle \\
    {\color{theoretical}Power} &  & \orangecircle & \orangecircle & \greencheck & \greencheck \\
    \midrule
    \multicolumn{6}{l}{\hyperref[sec:bayes_values]{\textit{Bayesian}}\& \hyperref[sec:kuhn_values]{\textit{Kuhnian}}} \\
    {\color{empirical}Accuracy} &  & \greencheck & \greencheck & \greencheck & \greencheck \\
    {\color{theoretical}Unification} &  & \redcross & \redcross & \redcross & \redcross \\
    \midrule
    \multicolumn{6}{l}{\hyperref[sec:kuhn_values]{\textit{Kuhnian}}} \\
    {\color{theoretical}Consistency} &  & \orangecircle & \greencheck & \greencheck & \greencheck \\
    {\textbf{\color{theoretical}Simplicity}} & \staricon & \orangecircle & \greencheck & \orangecircle & \greencheck \\
    {\textbf{\color{theoretical}Fruitfulness}} & \staricon & \orangecircle & \orangecircle & \redcross & \orangecircle \\
    \midrule
    \multicolumn{6}{l}{\hyperref[sec:deutsch_values]{\textit{Deutschian}}} \\
    {\textbf{\color{theoretical}Falsifiable}} & \exclamationicon & \greencheck & \greencheck & \greencheck & \greencheck \\
    {\textbf{\color{theoretical}Hard-to-vary}} & \staricon & \orangecircle & \orangecircle & \greencheck & \greencheck \\
    \midrule
    \multicolumn{6}{l}{\hyperref[sec:nomological_values]{\textit{Nomological}}} \\
    {\color{theoretical}Nomological} &  & \redcross & \redcross & \redcross & \orangecircle \\
    \bottomrule
    \end{tabular}
    \caption{An evaluation of MI explanation methods with respect to
    our Explanatory Virtues Framework as given in \Cref{sec:values}.
    The virtues which are indispensable for valid Mechanistic Interpretability explanations
    are highlighted with a \exclamationicon.
    The virtues that we consider to be the most important for good explanations
    are highlighted with a \staricon.
    Metrics are grouped by their philosophical foundations: Deutschian, Kuhnian, Bayesian, or Nomological.
    Blue metrics indicate empirical criteria, while orange metrics represent theoretical criteria.
    Green checks, orange circles and red crosses indicate that the method
    well-considers, moderately considers, or poorly considers the virtue, respectively.
    The explanatory case studies that we have considered
    generally optimise for accuracy, however
    they vary in their commitment to the virtues of Simplicity,
    Unification and Nomologicity.
    In our descriptions of these methods across \Cref{sec:explanations_in_the_wild},
    we provide a more detailed analysis of how we assess the virtues of each method
    and we provide our full evaluation rubric in \Cref{table:values_rubric}.
}
\label{table:values_case_study}
\end{table}

\section{The Road Ahead}
\label{sec:discussion}

\begin{quote}
    \itshape ``Science is built up of facts, as a house is built of stones;
    but an accumulation of facts is no more a science than a heap of stones is a house.''
    \par\raggedleft\normalfont --- \citet{poincare1905science}
\end{quote}

The field of Mechanistic Interpretability was founded by \citet{olah2020zoom_in_circuits} to distinguish
itself from previous approaches of neural network interpretability.
These previous approaches were not sufficiently grounded in causal abstraction, nor treated the model internals
appropriately as representing explanations as intrinsic structure that we would like to uncover
\citep{phil_exp_ayonrinde_jaburi, saphra2024mechanistic}.
The `Mechanistic turn' in interpretability was a step
towards unifying a community around faithful and falsifiable explanations of
models.
The Explanatory Virtues Framework is a further step in this direction,
providing unifying criteria to evaluate explanatory methods.
In particular, focusing on the following three virtues
would constitute methodological progress for the field:

\paragraph{1. Simplicity and Compression.}
\citet{Swinburne1997_Simplicity} argues that simplicity is a key virtue
of good explanations and can provide evidence to the truth of a theory.
However, appropriately characterising an explanatory Simplicity measure
is currently an open question for interpretability.\footnote{
    \citet{ayonrinde2024_mdl_saes,phil_exp_ayonrinde_jaburi}
    argue that good explanations in interpretability
    can primarily be understood as compressions of information.
}
Early explorations into understanding compression as a key function of explanation
can be found in
the Compact Proofs literature \citep{gross2024compact}
and Attribution-Based Parameter Decomposition \citep{braun2025apd}.
Coalescing around a concept of Simplicity for interpretability would allow
different explanations to be rigorously compared on the (accuracy, simplicity) Pareto curve,
which is directly useful in many applications.
Such a definition might also naturally encourage further research into the impact
of modularity in both neural networks and their explanations
\citep{clune2013modularity, filan2021clusterability, baldwin1999design_modularity}.

\paragraph{2. Unification and Co-Explanation.}
\citet{Hempel1966_philosophy_natural_science_unification}
argues that unification is a core driver of scientific progress.\footnote{
    See also \citet{kitcher1981explanatory_unification}.
}
Indeed we may see unification as a drive towards compression of explanations
where the set of phenomena to be explained is large
\citep{bassan2024local_global_interp, bhattacharjee2024auditing_local_explanations}.
Currently, most methods in interpretability don't seek to co-explain
many phenomena using the same building blocks.
The Mechanistic Interpretability (MI) community has sought to understand the universality (or otherwise)
of representations and algorithms across many models
with mixed results
\citep{olah2020zoom_in_circuits, olsson2022induction_heads, chughtai2023group_universality}.
However, we may also be interested in modular compositional explanations where the explanatory
units are shared not only across models but also across different tasks and domains within a
single model.
For example, there is evidence that induction heads are reused for many tasks
within models and so induction heads perform a co-explanatory function
\citep{olsson2022induction_heads}.

\paragraph{3. Nomological Principles.}
\citet{bacon1620novum} writes that any science first starts by observations.
After that point, most fields have a choice to make between two (non-exclusive) paths
that \citet{windelband1894geschichte_nomothetic_idiographic} refers to as the
\emph{nomothetic} and \emph{idiographic} approaches.
The nomothetic approach seeks to rapidly synthesise these early observations into
general explanatory theories with nomological principles that are useful
for making predictions.
Conversely, the idiographic approach focuses on categorising and describing
ever more exhaustive sets of observations, without necessarily seeking general laws
to explain them.\footnote{
    Physicist Ernest Rutherford famously (and somewhat disparagingly) referred to this
    difference as partitioning sciences into two cultures: physics and stamp collecting
    \citep{Bernal1940_rutherford_stamp_collecting_quote}.
}
Physics is a prototypical nomothetic science;
biology is often considered an idiographic science.\footnote{
    We might think of History as a highly idiographic field outside of the sciences.
}
Idiographic approaches can tend towards \emph{description} rather than \emph{explanation}.
For example, we might wonder if interpretability researchers counting up and
categorising all the features in a given model's latent space is much different
to a biologist naming and describing all the species of beetle in an ecosystem
without learning anything about the evolution of these species or how they
interact within the environment.

The use of nomological principles can simplify explanations
and help to provide a unifying paradigm for Mechanistic Interpretability.
Efforts in Developmental Interpretability \citep{hoogland2024dev_interp},
the Physics of Intelligence \citep{allenphysics_of_lm},
Computational Mechanics \citep{shai2024comp_mech},
and the Science of Deep Learning
\citep{lubana2023mechanistic_mode_connectivity,allen2023physics_of_lms}
may also produce useful nomological principles for the MI
community to adopt in their explanations.

Mechanistic Interpretability has begun to congeal into a genuine field,
with survey papers \citep{mech_interp_review_2024},
and a (major conference) workshop \citep{icml_mech_interp_workshop}.
There are principles, problems and methods that many members of the
Mechanistic Interpretability community adopt,
even if there are still many open questions and methodological disputes to address.
Though it is not strictly necessary to adopt a nomothetic approach for
a field to have a paradigm for Normal Science in the Kuhnian sense \citep{Kuhn1962-strcture_of_sci_rev},
nomothetically oriented fields with laws and principles to test and critique
have historically tended to see more rapid scientific progress.
We might regard the move towards a nomothetic and unifying approach
to explanations in Mechanistic Interpretability as a move towards a more mature
science. Efforts in the Science of Deep Learning seeking to develop principles
for neural networks may provide a basis for nomological principles
for Mechanistic Interpretability.

\citet{Shimi_2024_rocks} writes:
\emph{``At the beginning of every science there's a guy who's just cataloguing rocks.''}
We might add:
\emph{And then it turns out that we can use these observations to build a theory.
We should not be forever just cataloguing rocks;
we do actually have to build a theory at some point!}
Mechanistic Interpretability has found
Causal Abstractions theory to be a useful foundation.
We suggest that a further paradigm for Mechanistic Interpretability
should take seriously the virtues of good explanations.
The Explanatory Virtues allow us to iteratively build
better interpretability methods and
generate increasingly good explanations of neural networks.

Progress in Mechanistic Interpretability may provide insights into AI systems which
are useful for increasing the transparency and safety of systems which are
deployed widely and/or in critical applications
\citep{bengio2025ai_safety_report,reuel_military_risks,sharkey2025openproblemsmechanisticinterpretability}.
We believe that our Explanatory Virtues Framework can help researchers
in designing methods which lead to more reliable and useful explanations of neural systems.

\hypertarget{acknowledgments}{%
\subsubsection*{Acknowledgments}\label{acknowledgments}}
Thanks to Nora Belrose, Matthew Farr, Sean Trott, Evžen Wybitul, Andy Artiti,
Owen Parsons, Kristaps Kallaste and Egg Syntax for comments on early drafts.
Thanks to Elsie Jang, Alexander Gietelink Oldenziel, Jacob Pfau,
Catherine Fist, Lee Sharkey, Michael Pearce, Mel Andrews, Daniel Filan,
Jason Gross, Samuel Schindler, Dashiell Stander, Geoffrey Irving and
attendees of the ICML MechInterp Social for useful conversations.
We're grateful to Kwamina Orleans-Pobee, Will Kirby and Aliya Ahmad for additional support.
This project was supported in part by a Foresight Institute AI Safety Grant.

\section*{Reproducibility Statement}
The comparative evaluation of explanation methods presented in \Cref{table:values_case_study}
can be reproduced by applying the Explanatory Virtues Rubric detailed in \Cref{table:values_rubric}.
This rubric provides clear criteria for assessing the extent to which
different Mechanistic Interpretability methods embody each explanatory virtue.
By following the three-level assessment framework
(Highly Virtuous, Weakly Virtuous, Not Virtuous)
with their corresponding indicators (\greencheck, \orangecircle, \redcross),
researchers can systematically evaluate explanation methods against the Explanatory Virtues Framework.
The rubric's structured approach ensures that assessments are based on consistent
 criteria rather than subjective preferences,
 allowing for reproducible comparisons between different explanation
 methods in Mechanistic Interpretability.

\section*{Ethics Statement}
This work focuses on developing a philosophical framework for
evaluating explanations in the context of Mechanistic Interpretability of neural networks.
As a theoretical contribution, our framework itself does not directly raise ethical
concerns typically associated with empirical AI research,
such as data privacy, bias, or direct societal impacts.
However, we recognize that advances in Mechanistic
Interpretability have significant ethical implications.

Better explanations of AI systems, which our framework aims to encourage,
can promote transparency, accountability, and trust in AI systems.
We note that improved understanding of neural networks through Mechanistic Interpretability
may contribute to AI Safety, AI Ethics, and the responsible deployment
of AI systems in critical applications.
By providing systematic criteria for evaluating explanations,
our work supports the responsible development of AI that is interpretable
and human-understandable.

We hope this work contributes to the broader goal of developing AI
systems that can be meaningfully understood, monitored, and steered by humans.

\newpage

\bibliographystyle{colm2025_conference}
\bibliography{references}

\newpage
\appendix %

\newpage

\hypertarget{appendix_values}{%
\section{The Explanatory Virtues Rubric}\label{sec:values_rubric}}
\begin{longtable}{@{}p{2.2cm}p{3.8cm}p{3.8cm}p{3.8cm}@{}}
    \caption{
        The rubric for evaluating the Explanatory Virtues of a
        given explanation (see \Cref{fig:explanatory_values,sec:values}).
        We use this rubric to provide a structured evaluation of explanations
        as in \Cref{table:values_case_study}.}\\
    \toprule
    \textbf{Explanatory Virtue} & \textbf{Highly Virtuous} & \textbf{Weakly Virtuous} & \textbf{Not Virtuous} \\
    \specialrule{1.5pt}{3pt}{3pt}
    \endfirsthead

    \caption[]{The rubric for evaluating the Explanatory Virtues (continued)}\\
    \toprule
    \textbf{Explanatory Virtue} & \textbf{Highly Virtuous} & \textbf{Weakly Virtuous} & \textbf{Not Virtuous} \\
    \specialrule{1.5pt}{3pt}{3pt}
    \endhead

    \bottomrule
    \endfoot

        Icon &
            \greencheck &
            \orangecircle &
            \redcross \\
        \specialrule{1.5pt}{3pt}{3pt}

        {\color{theoretical}Causal-Mechanistic} &
            Generates end-to-end causal explanations &
            Explains a part of the network and can be used as part of a Causal-Mechanistic Explanation &
            Generates explanations which are not used for producing end-to-end causal explanations \\
        \specialrule{1.5pt}{3pt}{3pt}
        {\color{theoretical}Precision} &
            Rewards explanations that provide precise and risky predictions in a quantifiable way &
            Partially accounts for precision in explanations, possibly qualitatively &
            Fails to penalise (or even endorses) overly broad or vague predictions \\
        \midrule
        {\color{theoretical}Priors} &
            Explicitly incorporates comparisons with background theoretical priors in the method &
            Implicitly incorporates background theoretical priors in evaluating explanations &
            Fails to appropriately incorporate background theoretical priors \\
        \midrule
        {\color{empirical}Descriptiveness} &
            Prefers explanations which clearly analyse detailed, component-wise prediction quality in high fidelity, capturing the essential characteristics of each data point &
            Only partially tangentially analyses individual data point fit, mostly focusing on overall aggregated fit &
            No analysis of how the data points fit the explanation in isolation at all \\
        \midrule
        {\color{empirical}Co-Explanation} &
            Assesses the ability of explanations to account for multiple observations together, rewarding measures that emphasise integrated, joint predictive performance. &
            Has the potential to incorporate some aspects of joint explanation but does not fully reward coherent integration across diverse data points in its currently practised form &
             Evaluates each data point in isolation, ignoring the value of linking multiple observations. \\
        \midrule
        {\color{theoretical}Power} &
            Strongly values approaches that produce highly constraining predictions
            (especially about observations considered in isolation),
            penalising methods that allow too many plausible alternatives &
            Provides moderate emphasis on constraining predictions, allowing for some uncertainty. &
            Assigns no weight to the predictive force of the explanation \\
        \specialrule{1.5pt}{3pt}{3pt}
        {\color{empirical}Accuracy} &
            Quantitatively rewards explanations that fit the data with minimal error, especially does so with reference to both the precision and recall where relevant &
            Qualitatively rewards explanations that seem to fit the data well subjectively &
            Does not distinguish between explanations that fit the data well or poorly leading to evaluations that tolerate significant errors \\
        \midrule
        {\color{theoretical}Unification} &
            Measures how well a single evaluation framework can account for diverse observations, emphasizing integrated, unified explanations &
            Has the potential to recognise some unification even if in a limited or fragmented way or if this is not a typical application of the method &
            Places no weight on a unified account rather than a disjunction of accounts \\
        \specialrule{1.5pt}{3pt}{3pt}
        {\color{theoretical}Consistency} &
            Requires internal coherence within the explanation and multiple instances of running the same explanation method &
            Mostly internally consistent but probabilistically can provide inconsistent explanations &
            Places no weight on the internal consistency of generated explanations \\
        \midrule
        {\color{theoretical}Simplicity} &
            Evaluates explanations based on a conciseness or K-complexity simplicity metric rewarding simpler explanations &
            Partially considers a weak form of simplicity such as parsimony &
            Neglects simplicity as a factor, encouraging highly complex and complicated explanations \\
        \midrule
        {\color{theoretical}Fruitfulness} &
            Rewards explanations that predicted new, testable phenomena even with adversarially chosen test data from a close data distribution &
            Rewards explanations that predict novel phenomena even from the same data distribution &
            Assesses only current data fit with no train-val-test split at all \\
        \specialrule{1.5pt}{3pt}{3pt}
        {\color{theoretical}Falsifiable} &
            Requires that explanations yield clear, testable predictions and penalises those that could not be refuted under counterfactual data. &
            - &
            Fails to consider whether explanations can be empirically refuted, rewarding unfalsifiable evaluations. \\
        \midrule
        {\color{theoretical}Hard-to-vary} &
            Rigorously assesses the robustness of explanations, rewarding those evaluations where small modifications would lead to significant performance degradation. Checks for interdependencies among components to ensure that each part is essential and load-bearing &
            Makes limited effort to avoid ad-hoc explanations but doesn't fully address how hard-to-vary the explanations are &
            Does not account for the ease of altering explanations and consistently produces explanations that are easily tweaked without loss of predictive power \\
        \specialrule{1.5pt}{3pt}{3pt}
        {\color{theoretical}Nomological} &
            Explicitly integrates established general laws and principles, favouring evaluations that connect to a broader nomological framework or reusing laws in multiple places across the explanatory theory &
            Implicitly appeals to some non-generic laws but such a connection may be indirect and not well utilised &
            Ignores links to universal principles and attempts to focus on explaining the data without any reference to more general theoretical principles \\
    \label{table:values_rubric}
\end{longtable}

\hypertarget{straightforward_explanations}{%
\section{Straightforward explanations}\label{sec:straightforward_explanations}}

Following \citep{phil_exp_ayonrinde_jaburi}, we define the
\emph{straightforward explanation} of a neural network as follows.
Given a neural network $f:X\to Y$ and $x \in X$ such that  $f(x)=y$,
the straightforward explanation is given by the computational trace
of the network on the input x.\footnote{
    In fact, this explanation is a formal proof of the equality $f(x)=y$.
}
We note that for any neural network $f$ and sub-distribution $D \subseteq \mathcal{D}$,
there exists a straightforward explanation of $f$ on $D$.
However, this straightforward explanation is
typically not good a explanation in the sense of \Cref{sec:values} as
such explanations are not very concise or illuminating.
We would instead like explanations of neural networks that are in terms of the features
(or concepts) that the network learned during training and explanations which are compact and useful.

Given \Cref{sec:values} and \Cref{sec:values_rubric} we may evaluate the straightforward explanation
of a neural network using the Explanatory Virtues Framework.
    \begin{itemize}
        \setlength{\itemsep}{0pt}
        \item \textbf{Causal-Mechanistic:} The straightforward explanation is Causal-Mechanistic.
        It decomposes the model into a computational graph,
        given by the neural network architecture.
        \item \textbf{Precision, Descriptiveness, Accuracy, Power \& Falsifiable:} The straightforward explanation fulfills all these criteria,
        since it is a complete representation of the model.
        \item \textbf{Co-explanation \& Unification:} The straightforward explanation does not fulfill these criteria,
        since it treats all inputs independently.
        \item \textbf{Priors:} The straightforward explanation does not refer to priors in its interpretation.
        \item \textbf{Consistency:} The straightforward explanation is consistent.
        \item \textbf{Simplicity:} The straightforward explanation is highly complex.
        There is no compression from the original weights in the explanation given.
        \item \textbf{Fruitfulness:} The straightforward explanation is not fruitful,
        in that it doesn't provide novel predictions.
        \item \textbf{Hard-to-vary:} The straightforward explanation is not hard-to-vary;
        modifying single parts of the model (e.g. individual weights) by some small amount will typically not vary the model performance.
        \item \textbf{Nomological:} The straightforward explanation is not nomological as it doesn't provide general laws or principles.
    \end{itemize}

We note that the straightforward explanation is a valid explanation of a neural network:
It is Model-level, Ontic, Causal-Mechanistic, and Falsifiable.
Further, the straightforward explanation embodies many of the explanatory values.
However, we hope the reader will agree that the straightforward explanation is not a \emph{good explanation}.
Since, as noted in \Cref{sec:discussion}, not all of the explanatory values are equally as important,
an explanation may embody some of the virtues and yet not be a good explanation.

Researchers who are interpreting a neural network
may have different use cases for which they would like
an explanation of the model behaviour.
To account for these different goals,
researchers can make trade-offs between which Explanatory Virtues they
value most highly.\footnote{
Choosing the right explanation is a value-laden task \citep{phil_exp_ayonrinde_jaburi}.
}
Overall, however, for an explanation to be a good explanation,
we suggest that \emph{Simplicity} and
\emph{Fruitfulness} and \emph{Hard-to-Varyness} are the most
important values, without which it is difficult to have a good explanation.
In this case, the straightforward explanation fails on the virtue of Simplicity.

\hypertarget{explanations_figures}{%
\section{Explanations in The Wild, Visually}\label{sec:explanations_figures}}

This section is a visual companion to \Cref{sec:explanations_in_the_wild}.
We present a series of figures to elucidate what we mean by each form of
explanation and how we choose between two explanations given this method
(i.e. Theory Choice \citep{Schindler_theoretical_virtues}).

\subsection{Clustering (Activations or Inputs)}
\label{sec:clustering_fig}

\begin{figure}[H]
    \centering
    \begin{minipage}{0.8\linewidth}
        \centering
        \includegraphics[width=\linewidth]{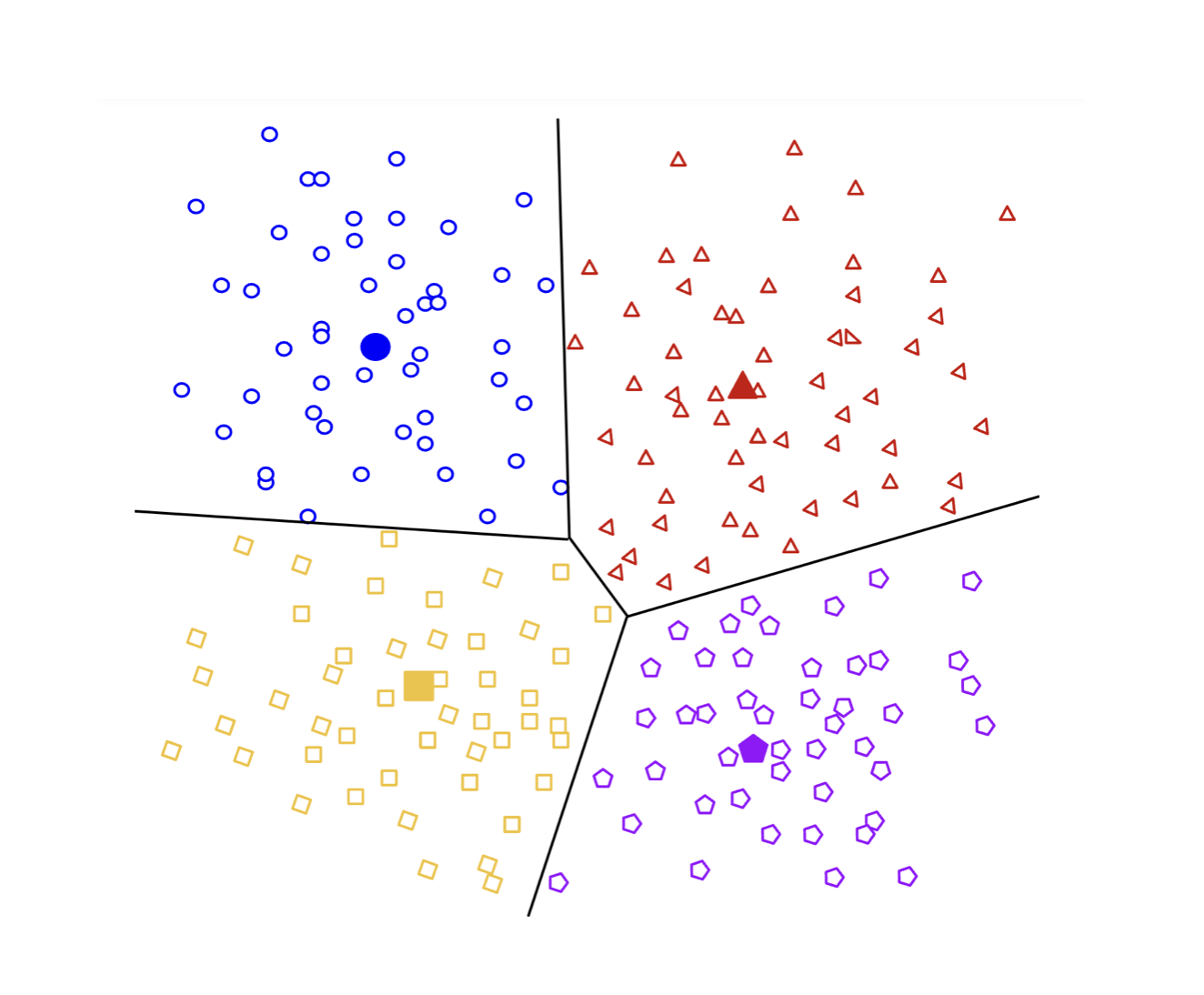}
    \end{minipage}
    \caption{Given some (possibly intermediate) embeddings ($\x$), a clustering explanation
    can be produced by assigning $\x$ to a cluster $C_i$, where the n clusters partition the
    input space into disjoint regions.
    Here $C_1 \cup C_2 \cup \ldots \cup C_n = \mathbb{R}^N$ and $C_i \cap C_j = \emptyset$  $\forall i \neq j$.
    The explanation is then given by taking the behaviour of the model on some cluster representative,
    or centroid, $\mu_i \in C_i$.
    We can intuitively see this as performing a quotient operation on the input space,
    where the model behaviour is approximated by a piecewise constant function.
    [Image from \citet{google_clustering}].
    }
\label{fig:clustering_explanations}
\end{figure}

\subsection{Sparse Autoencoder Explanations of Representations/Activations}
\label{sec:saes_fig}

\begin{figure}[H]
    \centering
    \begin{minipage}{0.9\linewidth}
        \centering
        \includegraphics[width=\linewidth]{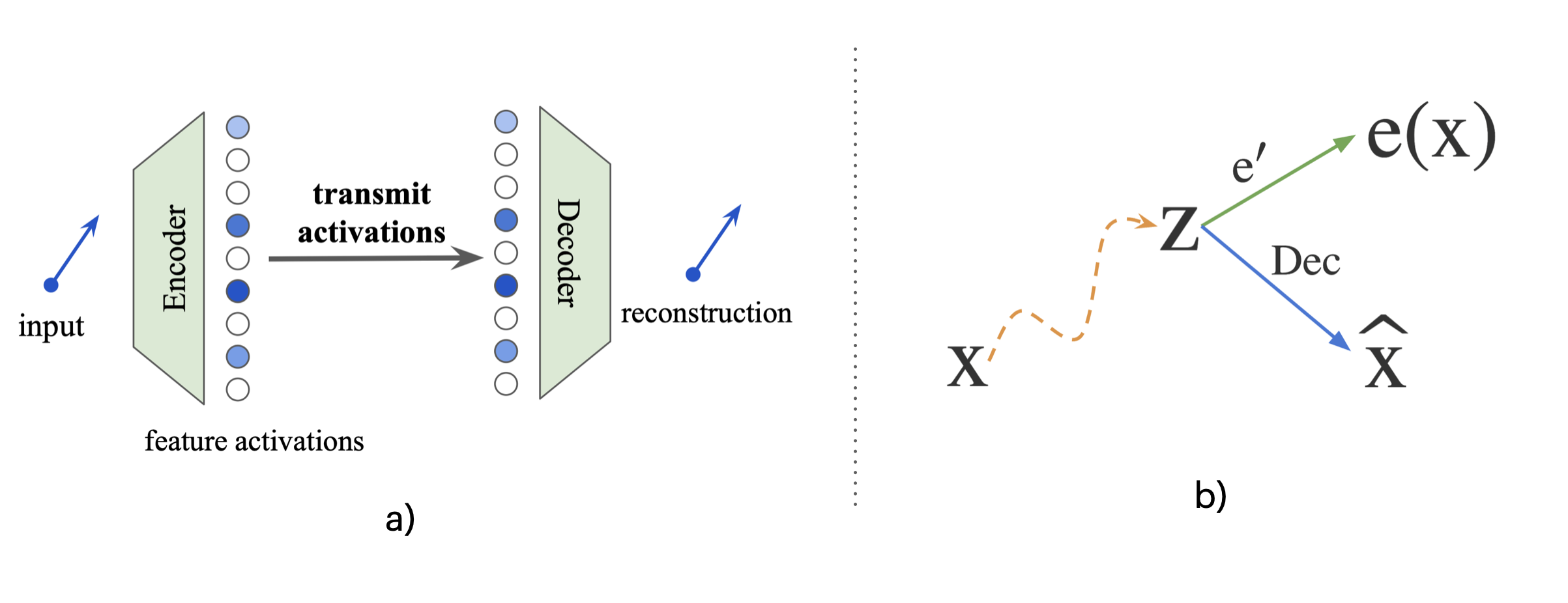}
    \end{minipage}
    \caption{
    (a) The SAE architecture. An encoder provides some set of latents (or feature activations) in the feature basis.
    We have some decoder map, Dec, which is a linear combination of the columns of the feature dictionary weighted by the sparse latents.
    We say informally that these latents \emph{correspond} to the input activations if, under the decoder
    map, Dec.
    (b) If $\x$ and $\z$ correspond in the above sense then
    the natural language explanation of the input activations $\x$
    is given as $e(\x) = e'(\z)$; that is the explanation of the latents
    using the automated interpretability process $e'(\z)$
    \citep{paulo2024autointerp,saebench2024,bills2023autointerp,kayonrinde2025incoherent_saes}.
    We can measure the mathematical description length (\emph{Conciseness}) of the explanation $e(\x)$
    as the number of bits required to describe the latents $\z$ \citep{ayonrinde2024_mdl_saes}.
    [Images from \citet{ayonrinde2024_mdl_saes,kayonrinde2025incoherent_saes}]
    }
\label{fig:sae_explanations}
\end{figure}

\subsection{Causal Abstraction Explanations of Circuits}
\label{sec:circuits_fig}

\begin{figure}[H]
    \centering
    \begin{minipage}{0.9\linewidth}
        \centering
        \includegraphics[width=\linewidth]{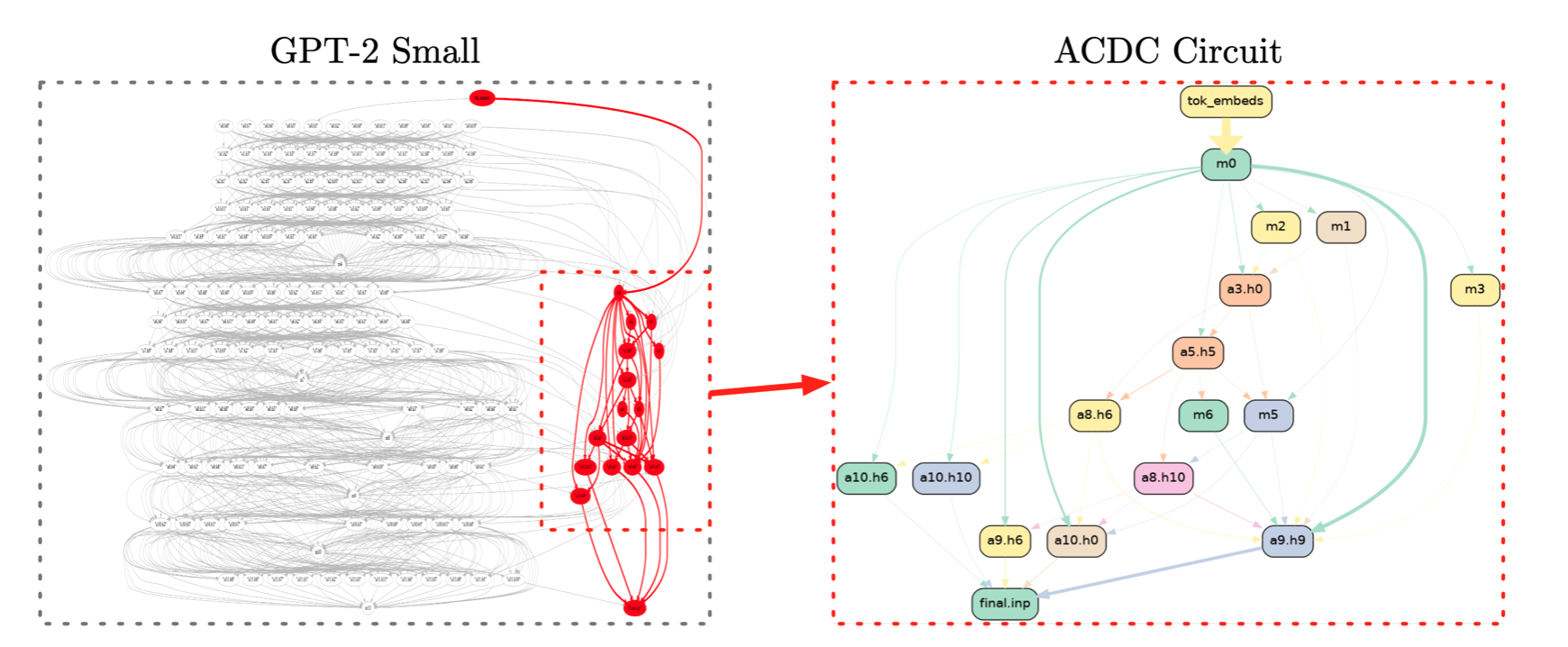}
    \end{minipage}
    \caption{A circuit explanation is a Causal-Mechanistic explanation
    such that the circuit C is a constructive abstraction of a neural network's computational graph
    M if there exists a partition the variables in M such that each high-level variables in C correspond to a
    low-level partition cell in M and interventions on M correspond to interventions on C.
    For example in \Cref{fig:circuits_acdc} \emph{Left} \citep{conmy2023acdc},
    the IOI circuit \citep{wang_ioi} (highlighted in red) is recovered from the computational graph
    of GPT-2 Small.
    [Image from \citep{conmy2023acdc}].
    }
\label{fig:circuits_acdc}
\end{figure}

\subsection{Compact Proofs}
\label{sec:compact_proofs_fig}

\begin{figure}[H]
    \centering
    \begin{minipage}{1.0\linewidth}
        \centering
        \includegraphics[width=\linewidth]{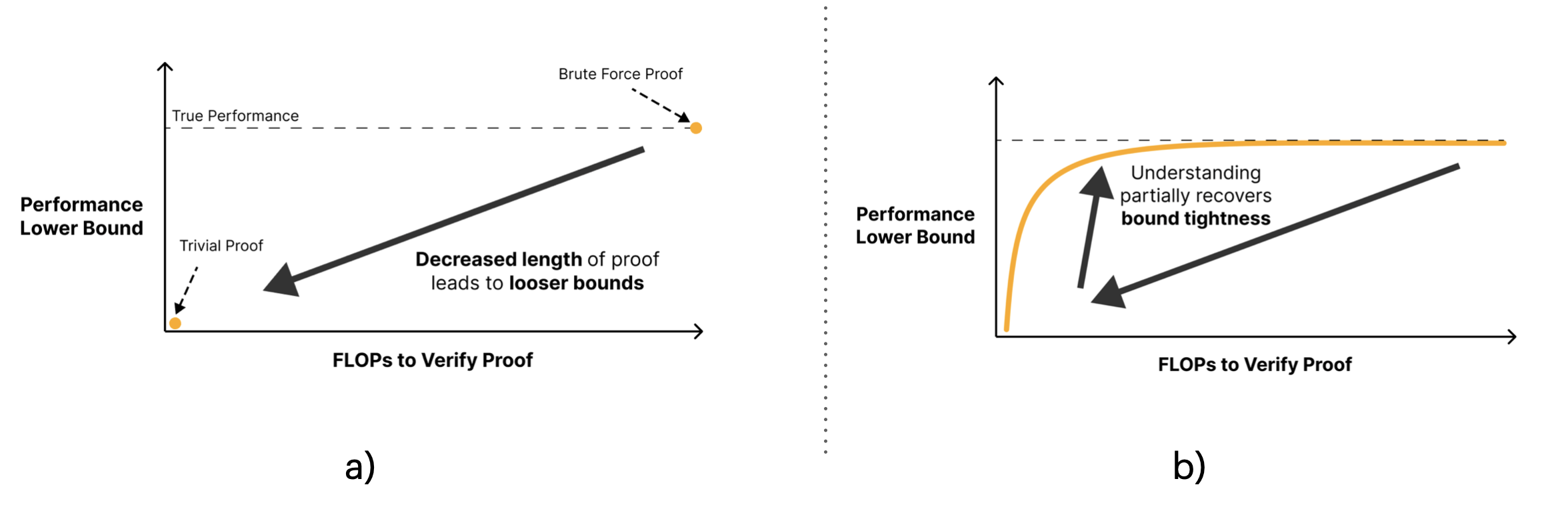}
    \end{minipage}
    \caption{(a) Compact Proofs evaluate explanations on two metrics,
    their compactness (FLOPs to Verify Proof)
    and their accuracy (Model Performance Lower Bound).
    These two metrics can be assessed on a Pareto frontier.
    (b) A good explanation should push the frontier towards the upper left corner (i.e. more accurate and compact proofs).
    [Image from \citet{gross2024compact}.]
    }
\label{fig:compact_proofs}
\end{figure}

\hypertarget{appendix_examples}{%
\section{Examples of Explanations}\label{sec:examples}}

In this section, we provide some intuitive examples and non-examples of Explanations
which satisfy the criteria that we outline above.
The case studies in \Cref{sec:explanations_in_the_wild} are examples
within Mechanistic Interpretability and Machine Learning;
our examples here are non-technical illustrations.

\subsection{Examples of Explanation Types}
\label{sec:example_explanation_types}

\subsubsection{Ontic Explanations}

\emph{Question}: Why did the pen fall off the desk?

\paragraph{Causal-Mechanistic But Not Ontic Explanation.}
\begin{quote}The pen fell off the desk because the aether pushed the bottle and then the bottle pushed the pen off the desk.
\end{quote}
This explanation is Causal-Mechanistic in the sense that one thing happens after another and causes the next.
However, if we do not believe that the aether is a real entity then this explanation cannot be considered an Ontic Explanation.

---

\emph{Question}: Why is the cube heavy?

\paragraph{Ontic But Not Causal-Mechanistic Explanation.}
\begin{quote}The cube is heavy because it is made up of tungsten atoms. \end{quote}
This explanation is Ontic as the entities involved in the explanation are
real entities.
However, it is not Causal-Mechanistic as there is no step-by-step explanation without gaps.

\subsubsection{Statistically-Relevant Explanations}

Consider the explanation:
\begin{quote}
    Ice cream sales are higher on days when there are more shark attacks.
    If there's a shark attack reported,
    we can predict with 85\% confidence that ice cream sales will be above average that day.
\end{quote}
This explanation is purely in terms of statistical correlation rather than causation.
There is no explication of any underlying causal mechanism,
which might involve both phenomena being causally downstream of hot
weather and/or more beach visitors.
We could perform interventions to test this hypothesis.

\subsubsection{Nomological Explanations}

\emph{Question}: Why does a metal rod expand when heated?

\paragraph{Nomological but not Causal-Mechanistic explanation.}
\begin{quote}
    The rod expands because it follows the natural law that all metals expand when heated,
    as described by the coefficient of thermal expansion.
\end{quote}
This explanation references a general law of nature without getting into the underlying mechanism.

\paragraph{Causal-Mechanistic Explanation.}
\begin{quote}
    The rod expands because
    its metal atoms vibrate more vigorously when heated, which increases their average spacing.
    This increased spacing leads to an overall increase in the rod's length.
\end{quote}
This details the physical mechanism causing the expansion.

\subsection{Examples of Explanatory Values}
\label{sec:examples_for_explanatory_values}

\subsubsection{Precision, Power and Unification}

Consider one explanation of what happens to objects when they are dropped:
\begin{quote}
    When an object is dropped, it falls to the ground due to the force of gravity.
\end{quote}
compared to the more \textbf{precise} explanation:
\begin{quote}
    Objects fall toward Earth at a rate of 9.8 meters per second squared,
    with slight variations depending on altitude and latitude.
\end{quote}

The latter explanation rules out more possibilities than the former.
When we see that an object is dropped, armed with the second explanation,
we are able to rule out the possibility that the object will fall at a different rate
as well as the possibility that it will rise into the air.

Precise explanations make \emph{narrow} and \emph{risky} predictions.

\paragraph{Unification.} An explanation is \textbf{unifying} if it purports to explain multiple disparate observations.
The Central Dogma in molecular biology states that genetic information flows only in one direction,
from DNA, to RNA, to protein, or RNA directly to protein \citep{crick1970central_dogma}.
This theory operates as a unifying explanation which narrows the space of possibilities
for a wide range of biological phenomena.

\subsubsection{Consistency}

Consistent explanations contain no internal contradictions.

\emph{Question}: Why did Alice miss the important meeting this morning?

\paragraph{Inconsistent Explanation.}
Alice, being a forgetful person,
forgot that the meeting was happening and simultaneously Alice
deliberately skipped the meeting to avoid a confrontation.

\paragraph{Consistent Explanation.}
Alice was out of the office for a vacation and missed the meeting.

As we increase the unification/scope of explanations, we sometimes introduce inconsistencies.
For example, as we look to unify Quantum Mechanics and General Relativity,
two explanations which are internally consistent on their own,
we find that they are inconsistent with each other.

\subsubsection{Simplicity}

Occam's Razor states that when faced with competing explanations,
one should select the explanation that is the simplest.
This heuristic was first formulated in terms of parsimony,
but we might also extend the sense of simplicity here to
conciseness (Shannon complexity) or K-complexity (Kolmogorov complexity)
as more appropriate measures of simplicity.

The Ptolemaic explanation:
\begin{quote}
    The Earth is at the center of the universe,
    with the planets, the sun, and stars orbiting around Earth.
    There are many epicycles which explain the retrograde motion of the planets
    (planets moving backwards in the sky).
\end{quote}
is more complex than the Copernican explanation:
\begin{quote}
    The sun is at the center of the solar system
    and the planets orbit the sun in ellipses.
\end{quote}

Even though both explanations could fit the data,
we ought to prefer the Copernican model
according to Occam's Razor and
our Explanatory Virtue of Simplicity.

\citet{dedeo_2020_probability_consilience} give a sobering example of
the dangers of not sufficiently valuing simplicity in explanation
in their analysis of conspiracy theories.
Such theories are often ``abnormally co-explanatory and descriptive \ldots,
account for anomalous facts which are unlikely under the `official' explanation \ldots,
show how seemingly arbitrary facts of ordinary life are correlated by hidden events \ldots,
and describe a unified universe where everything is correlated by a network of hidden common
causes.''
A primary reason that such conspiracy theories are not typically good explanations
is that they are not \emph{simple}:
there's often a large amount of complexity and ad-hoc reasoning
to explain contradictory evidence and the reason for why the cover-up
has yet to come to light.

\subsection{Falsifiability and Hard-To-Varyness}

\citet{popper_scientific_discovery} argues against the pseudoscientific
theories of Marx, Freud, and Adler on the grounds that they are not falsifiable.
That is to say, there exists no observation that could be made that would
contradict the theory and cause its proponents to abandon it.
For a theory to be falsifiable it must make some concrete predictions
about the world that could in principle be tested.

Consider the following three explanations for why there are seasons
(adapted from \citet{deutsch2011beginning_of_infinity}):

\paragraph{Not Falsifiable.}
\begin{quote}The seasons change when Zeus feels like it.\end{quote}

This explanation is not falsifiable because it does not make any predictions.
If there were no seasons one year, then it would not be a mark against the theory.

\paragraph{Falsifiable but Not Hard-To-Vary.}
\begin{quote} Demeter (the Greek Goddess) negotiates a deal with Hades such that her daughter Persephone
    visits Hades once a year. When Persephone is with Hades and not with her mother, Demeter is
    sad and the world becomes cold.\end{quote}

This explanation does make a concrete prediction: the seasons will change exactly once a year.
Another prediction that follows is that winter
(the period of cold where Persephone is with Hades) should happen
everywhere on Earth at the same time.
This explanation is falsified by the fact that the seasons are at different times in
Australia to in Athens.
The explanation is not very Hard-to-Vary however.
We could easily change any of the characters or mechanisms involved in the theory
and keep the same predictions.

\paragraph{Falsifiable and Hard-To-Vary.}
\begin{quote}The Earth’s axis of rotation is tilted relative to the plane of its orbit around the
sun. Hence for half of each year the northern hemisphere is tilted
towards the sun while the southern hemisphere is tilted away, and for
the other half it is the other way around. Whenever the sun’s rays are
falling vertically in one hemisphere (thus providing more heat per
unit area of the surface) they are falling obliquely in the other (thus
providing less heat). \end{quote}

This explanation is both falsifiable and hard-to-vary.
All of the details of the theory play a functional role
and cannot be easily changed.
The axis-tilt theory also (correctly) predicts the fact that
the seasons are reversed in the northern and southern hemispheres.

\subsection{(Mundane) Accuracy and Fruitfulness (Novel Success)}

Explanations have Mundane Accuracy insofar as they correctly account for the phenomena they aim to explain.
Conversely explanations are Fruitful if they predict new phenomena that were not available to the explainer at
the time of coming up with the explanation.
Being able to predict and explain new, previously unobserved phenomena that are later confirmed (as in Fruitfulness)
is typically considered more valuable than merely explaining known phenomena (as in Mundane Accuracy).

Einstein's General Relativity predicted that light would bend around massive objects like the sun
\citep{einstein1916relativity}.
In 1919, during a solar eclipse, Arthur Eddington observed that starlight passing near
the sun was indeed deflected by precisely the amount Einstein had predicted
\citep{dyson1920eddington_experiment_eclipse, kennefick2021confirming_einstein}.
Given that the phenomenon of light bending around massive objects was previously unknown,
this was a novel empirical success for Einstein's theory.
This can increase our credence in Einstein's theory because the prediction
was made before the observation, was precise and quantitative in an unknown domain
and the observations matched the prediction with high accuracy.

\subsection{Co-Explanation and Descriptiveness}

Explanations can be purely \emph{descriptive},
in which case they account well for the phenomena they aim to explain
but do not connect with other explanations.
Alternatively, explanations can be \emph{co-explanatory},
unifying phenomena that were previously thought to be distinct.

\paragraph{Descriptive but Not Co-Explanatory.}
\begin{quote}
Electricity involves the movement of charges and produces effects such as static attraction,
lightning, and electrical current.
Magnetism, on the other hand,
involves the attraction or repulsion between certain materials like lodestone and iron,
and manifests in the behavior of compasses pointing north.
\end{quote}

\paragraph{Co-Explanatory.}
\begin{quote}
Electricity and magnetism are manifestations of a single underlying electromagnetic force.
A changing electric field produces a magnetic field, and a changing magnetic field produces an electric field.
Moving electric charges create magnetic fields, while moving magnets induce electric currents.
\end{quote}

\hypertarget{appendix_adhoc}{%
\section{A Coherence Formulation of Adhocness}\label{sec:adhocness}}

\citep{Schindler_theoretical_virtues} also gives an \textcolor{blue}{adhocness} test for explanations which can
identify those which are the result of a post-hoc epicycle added to an easy-to-vary explanation.
For Schindler, an explanation is \textcolor{blue}{adhoc} if the modification $\Delta$
which it corresponds to is some additional hypothesis H (which we can think
of as being added in order to accommodate some awkward-to-explain data $\x_I$) and two conditions are met:
\begin{enumerate}
    \item H explains $\x_I$. That is $\Prob(\x_I | E, H) > \Prob(\x_I | E)$.
    \item Neither the original explanation E nor background theories B give evidence for H.
    That is $ \Prob(H | E, B)< \Prob(H)$.
\end{enumerate}

We define an \textcolor{blue}{adhocness} metric as Adhoc $= \Prob(H) - \Prob(H | E, B)$
where larger ad-hocness values are more adhoc and dispreferred.

\hypertarget{appendix_local_decodability}{%
\section{Local Decodability as an Explanatory Virtue}\label{sec:local_decodability}}

Another virtue that we may consider for highly unifying explanations is \textcolor{blue}{local-decodability}.
Locally decodable explanations allow for retrieval and use of some small segment of the explanation
without querying the whole explanation, analogously to locally-decodable error-correcting codes
\citep{yekhanin2012locally_decodable_codes}.
This is important as we would like not only for our explanations to
have information compression (concise representation) but
also information accessibility (the ability to retrieve specific subparts quickly).
In practice, an explanation of network which is compressed but not \textcolor{blue}{locally-decodable}
requires significant computational resources to query and is not useful for human
understanding.\footnote{
    Local decodability is measured in query complexity:
    the number of queries required to recover 1 bit of the message (explanation).
    Conciseness and query complexity are known to be inversely proportional but the
    exact fundamental limit on their relationship is currently unknown.
}
The Independent Additivity condition from \citet{ayonrinde2024_mdl_saes} is an example of a \textcolor{blue}{local-decodability} condition
in Mechanistic Interpretability.
V-Information \citep{Xu2020v_information} provides a useful
analogy for local-decodability in Machine Learning.

\end{document}